\documentclass[letterpaper, 10 pt, conference]{ieeeconf}  
\usepackage{xcolor}

\makeatletter
\let\NAT@parse\undefined
\makeatother
\usepackage[colorlinks,linkcolor=blue,citecolor=green]{hyperref}
\usepackage{amsmath,amssymb,amsfonts}
\usepackage{algorithmic}
\usepackage{graphicx}
\usepackage{textcomp}
\usepackage{wrapfig}
\usepackage[numbers,sort&compress]{natbib}
\usepackage{amsmath}
\usepackage{float}
\usepackage{bm}

\usepackage{overpic}
\usepackage[justification = centering,labelsep = period]{caption}
\usepackage{algorithm}
\usepackage{array}
\usepackage[caption=false,font=normalsize,labelfont=sf,textfont=sf]{subfig}
\usepackage{stfloats}
\usepackage{url}
\usepackage{verbatim}

\usepackage{color}
\usepackage{booktabs}
\usepackage{multirow}
\usepackage{makecell}
\usepackage{tabularx}
\usepackage{cleveref}
\crefname{figure}{Fig.}{Figs}  
\Crefname{figure}{Figure}{Figures}  
\crefname{table}{Table}{Tables}  
\Crefname{table}{Table}{Tables}  

\IEEEoverridecommandlockouts                              

\title{\LARGE \bf
Joint Angle Estimation with Customized Wristband Based on Online Incremental Learning
}

\author{Shuo Wang$^{1}$, Xiaobin Chen$^{1}$ and Xiaoming Tao$^{1}$
\thanks{$^{1}$Shuo Wang, Xiaobin Chen and Xiaoming Tao are with Research Institute for Intelligent Wearable Systems, The Hong Kong Polytechnic University, Kowloon, Hong Kong.}%
}

\begin{document}

\maketitle
\thispagestyle{empty}
\pagestyle{empty}

\begin{abstract}

Intelligent wearable technology plays an increasingly important role in human-computer interaction, motion, and health monitoring. To ensure comfort and practicality of use, one common form for motion monitoring is to utilize soft wearable sensors. However, many research applications regarding wearable sensors are simplistic and difficult to adapt to different situations. This study proposes a system for estimating the angle of the wrist joint using a customized wristband based on an online incremental learning approach. It is a two-stage estimation method: the first stage updates the model based on the wearer's wrist movement characteristics using online learning, integrating real-time data from an IMU as ground truth. The second stage utilizes the updated model for estimation of wrist joint angle solely with the wristband. In other words, model training is completed during data acquisition, allowing the trained model to be used for subsequent angle estimation. This method offers advantages in adapting to data drift caused by variations in different testing configurations, such as the left and right wrists of the same subject, deviations in the wearing position on the same wrist, and even differences among various subjects. The results indicate that the sensors exhibit good performance under strain variations, and the wrist joint trajectory estimation of the proposed system has an approximate error of 15$^\circ$ in different scenarios.

\end{abstract}

\section{INTRODUCTION}

Wearable electronic systems have shown great potential in human-computer interaction \cite{1,2,3}, human motion detection \cite{4,5,6} and healthcare scenarios \cite{7,8,9}. For human motion detection, rigid wearable devices, due to their rigidity and lack of skin conformity, often cause inconvenience and discomfort for the wearer \cite{10}. Wearable systems composed of flexible sensors, especially those based on fabric, are gaining increasing attention due to their softness, breathability, and ease of integration into clothing \cite{11,12,13,14,15,16}, which can achieve diverse functions through the collaboration of appropriate data collection and analysis systems.

Different kinds of fabric-based wearable systems have been developed such as the intelligent glove \cite{17} used for hand gesture recognition and the wristband \cite{18,19} used for hand motion monitoring and healthcare. However, most of the research \cite{20,21,22,23} involves simply developing a sensor and using machine learning or deep learning methods for offline signal classification, which often does not address the data drift encountered during actual use, rendering the recognition algorithms ineffective. Here, we present a customized wristband and an online incremental learning method for real-time  estimation of wrist joint angle. The customized wristband together with the online incremental learning is a novel exploration of wrist joint angle estimation using fabric wearable systems, which first ensures the comfort of wearing, and secondly, it provides a certain adaptability to data drift without the need for repeated offline data collection and training, making the entire estimation process easier to operate.

This work is based on the fabric sensors we previously developed, and then we design and fabricate a wristband for wrist joint angle estimation. We detail the fabrication process of the fabric sensing wristband and demonstrate the wristband's ability to estimate changes in wrist joint angles during wrist movements. The online incremental learning is introduced for real-time data processing. This work seeks to demonstrate the possibility of using the online learning method for the real-time application of wearable systems.

\section{DESIGN AND FABRICATION OF THE SMART WRISTBAND}
\label{math model of strain sensors}
\begin{figure*}[htp]
\centering
\vspace{-0.2cm}
\includegraphics[width=\textwidth]{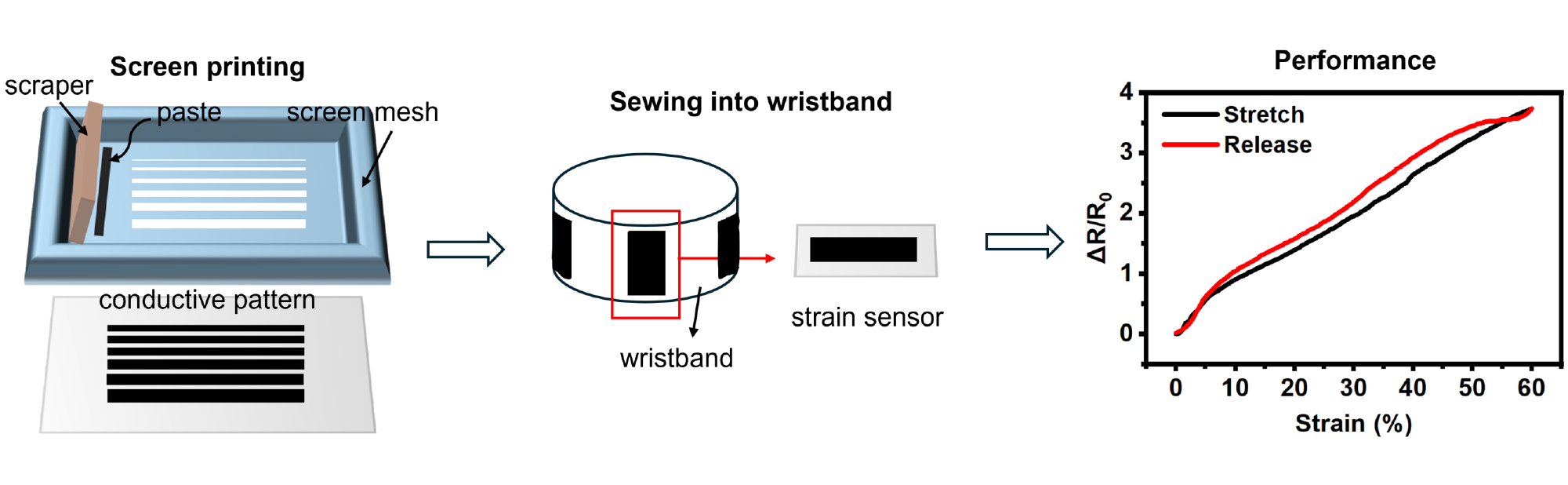}
\caption{Fabrication of the fabric strain sensor and smart wristband.}
\label{Fabrication}
\vspace{-0.6cm}
\end{figure*}

Resistive strain sensors are capable of converting mechanical deformation into resistance signals, enabling the detection of objects' strain state. Their core structure typically consists of flexible substrate materials (such as textile or polyimide) and conductive sensitive materials (such as metal nanowires or carbon composite). When subjected to external strain and deformation, the microstructure (the length of the conductive path or number of contact points) of the conductive material undergoes changes, leading to a corresponding change in resistance. Based on our previous work \cite{chen2022single, chen2024novel}, the fabric strain sensor was used to fabricate a smart wristband. The fabrication progress is shown in \cref{Fabrication}. Firstly, carbon black (CB) was selected as the conductive filler, room temperature vulcanized silicone elastomer (SE) was used as the elastomer matrix and nontoxic dimethyl silicone oil (SO) was used as a solvent. These three materials were mixed in a weight ratio of 1:10:15 and stirred thoroughly to obtain a carbon composite conductive material. Subsequently, the composite material was coated onto the plain knitted fabric substrate through manual screen printing and the strain sensors were obtained after drying. Finally, four sensors were cut and fixed with metal button at both ends serving as external electrical connections. These sensors were individually sewn into the wristband in the front, rear, left, and right directions to obtain the smart wristband. 

To acquire the resistance value of the sensors, a simple voltage divider circuit was built. One end of each sensor was connected to electrical ground (GND), and the other end was connected in series with a  resistor with 50$\ k\Omega$ and then connected to $VCC$ to form a voltage divider circuit. The resistor of the sensor can be calculated by measuring the voltage between the sensor and the variable resistor according to \eqref{res}:
\begin{equation}
\label{res}
V_{adc} =  VCC\times\frac{R_{s}}{R_{s}+R_{f}}
\end{equation}
Where $VCC$ is the input voltage of 3.3 V, $R_{s}$ is the resistance value of the sensor, $R_{f}$ is the reference resistance value, and $V_{adc}$ is the divided voltage value of the sensor, read by the ADC (Analog-to-Digital Converter) of microcontroller Arduino Uno. 

The performance of the sensor was evaluated by monitoring changes in resistance during tensile displacement. As shown in \cref{Fabrication}, the gauge factor (GF) values were computed as 10.3 for strains in 0\% - 6\% and 3.3 for strains ranging in 6\% - 60\%, indicating that the sensor not only operates effectively in high-strain regions far exceeding the limits of conventional metal strain gauges but also exhibits high sensitivity for detecting minor deformations. Furthermore, the resistance of the sensor returned to its initial state after the tensile force was removed, showing its excellent recovery properties and applicability in wearable applications.

\section{ONLINE INCREMENTAL LEARNING-BASED METHOD}
\subsection{Task Definition}
As mentioned in section \ref{math model of strain sensors}, when sensors are closed to human skin and deformed due to changes in wrist joint angle, there is a mapping from the strain space to the angle space of the wrist joint, which can be denoted as \eqref{deqn_ex1}.
\begin{equation}
\label{deqn_ex1}
\theta =  g(\epsilon)
\end{equation}

Where $\theta \in \mathbb{R}^3$ represents wrist joint angle obtained from the attached IMU, which will be detailed introduced in section \ref{Experiment Setup and Process}, $\epsilon$ is the strain values of the strain sensors and $g$ is the mapping function.

Because it is difficult for just one sensor to represent the wrist joint's angular changes accurately, as it would for a single degree-of-freedom joint, four sensors are used in this task. On the other hand, determining the appropriate calibration for each sensor's proportional coefficient and bias is a time-consuming process. An easier way is to direct map the resistance to the wrist joint angle directly. Meanwhile, resistance can be inferred from the voltage divider relationship with a reference resistor in the sensing circuit. Then, the direct mapping from voltage readings from circuit to wrist joint angle can be denoted as $g_{direct}$, which is shown as \eqref{deqn_ex2}.
\begin{equation}
\label{deqn_ex2}
\theta = g_{direct}(V_{1}, V_{2}, \ldots V_{m})
\end{equation}
Where $m$ is the number of strain sensors and $V_{i}$ represents the voltage reading of the $i_{th}$ strain sensor.

In previous research, diverse neural networks based on backward propagation will be utilized for the previous mentioned mapping. However, this approach requires large amount of data and time for offline training for each specific individual and will not perform well when data drift occurs. Therefore, we propose to utilize the online incremental learning method \cite{wang2023learning,fang2019vision} in this scenario, which mainly contains a  \textbf{Online Sequential Extreme Learning Machine (OSELM)} method for wrist joint angle estimation and a \textbf{Particle Swarm Optimization (PSO)} method for the initialization of parameters of the OSELM. 

\subsection{Initialization with PSO}
PSO is a widely utilized algorithm in various applications due to its ease of implementation and strong global search capabilities. \cite{daviran2025optimized, adnan2021improving}. Here, PSO is utilized for the number decision of the hidden nodes for ELM. 

\subsection{Online Sequential Extreme Learning Machine for Data Processing of IMU and Strain Sensors}
For the $i_{th}$ readings of $n$ strain sensors on the wristband from the readout circuit, they can be regarded as inputs $R_{i} = [r_{i1}\;r_{i2}\; ...\; r_{in}]^{\mathsf{T}}$ of an ELM. The outputs can be obtained with an IMU worn on the hand denoted as $\theta_{i}$. The mapping 
\begin{equation}
    \theta_{i} = f(R_{i})
\end{equation}
is to be learned. With the collected $N_{s}$ groups of samples, the output matrix of the hidden layer can be represented as follows:
\begin{equation}
\label{deqn_ex4}
\mathbf{H}
= 
\begin{bmatrix}
\Phi(a_{1},b_{1},R_{1}) \ldots \Phi(a_{N},b_{N},R_{1}) \\
\vdots\\
\Phi(a_{1},b_{1},R_{N_{s}}) \ldots \Phi(a_{N},b_{N},R_{N_{s}})
\end{bmatrix}_{N_{s}\times N} 
\end{equation}
Where $a$ and $b$ are weights and biases of hidden nodes, and $\Phi$ is the activation function. $N$ is the number of hidden nodes.

Then the output of a SLFN with $N$ hidden nodes can be shown as \cite{huang2011extreme}:
\begin{equation}
    \mathbf{H}\bm{\beta} = \bm{\theta} 
\end{equation}
Where $\bm{\beta} = [\beta_{1}^\mathsf{T}\; \beta_{2}^\mathsf{T}\; \ldots \; \beta_{N}^\mathsf{T}]$  is the weighting vectors of hidden nodes to the output nodes.

The objective is represented as below:
\begin{equation}
    \bm{\beta} = \mathop{\arg\min}\limits_{\bm{\beta}}\Vert \mathbf{H}\bm{\beta} -\bm{\theta} \Vert \
\end{equation}
This objective function can be solved with matrix operations and the result can be solved as:
\begin{equation}
    \bm{\beta_{solve}} = \bm{H}^\dagger\bm{\theta}
\end{equation}
Where $\bm{H}^\dagger$ is pseudo-inverse of $\bm{H}$.
Then, the first $\bm{H}^{(0)}$ can be calculated according to the first $N_{0}$ groups of samples, and the corresponding $\bm{\beta_{0}}$ can be obtained as:
\begin{equation}
\label{deqn_ex8}
    \bm{\beta}^{(0)} = \left(\bm{K}^{(0)}\right)^{-1}\left(\bm{H}^{(0)}\right)^{\mathsf{T}}\bm{\theta}^{(0)}
\end{equation}
Where $\bm{K}^{(0)} = \left(\bm{H}^{(0)}\right)^{\mathsf{T}}\bm{H}^{(0)}$, and when there are new $N_{1}$ samples come with the data stream, the corresponding $\bm{\beta}^{(1)}$ can be obtained as:
\begin{equation}
\label{deqn_ex9}
    \bm{\beta}^{(1)} = \left(\bm{K}^{(1)}\right)^{-1}
    \begin{bmatrix}
        \bm{H}^{(0)}\\
        \bm{H}^{(1)}
    \end{bmatrix}^{\mathsf{T}}
    \begin{bmatrix}
        \bm{\theta}^{(0)}\\
        \bm{\theta}^{(1)}
    \end{bmatrix}
\end{equation}
where $\bm{K}^{(1)} = \begin{bmatrix}
    \bm{H}^{(0)}\\
    \bm{H}^{(1)}
\end{bmatrix}^{\mathsf{T}}
\begin{bmatrix}
    \bm{H}^{(0)}\\
    \bm{H}^{(1)}
\end{bmatrix}$.
\\
\\
According to \eqref{deqn_ex8} and \eqref{deqn_ex9}, The recursive relation for $\bm{K}$ and $\bm{\beta}$ can be obtained as:
\begin{equation}
\label{eq10}
    \bm{K}^{(k+1)} = \bm{K}^{(k)} + \left(\bm{H}^{(k+1)}\right)^{\mathsf{T}}\bm{H}^{(k+1)}
\end{equation}

\begin{align}
\label{eq11}
    \bm{\beta}^{(k+1)} = \bm{\beta}^{(k)} + \left(\bm{K}^{(k+1)}\right)^{-1}\left(\bm{H}^{(k+1)}\right)^{\mathsf{T}} \notag \\ \left(\bm{\theta}^{(k+1)}-\bm{H}^{(k+1)}\bm{\beta}^{(k)}\right)
\end{align}

and with \eqref{eq10}, \eqref{eq11} and $Woodbury$ formula, when $\bm{M}^{(k)} = \left(\bm{K}^{(k)}\right)^{-1}$, the recursive relation can be further obtained as:
\begin{align}
    \bm{M}^{(k+1)} = \bm{M}^{(k)} - \bm{M}^{(k)}\left(\bm{H}^{(k+1)}\right)^{\mathsf{T}}
    \notag \\ 
    \left(\bm{I} + \bm{H}^{(k+1)}\bm{M}^{(k)}\left(\bm{H}^{(k+1)}\right)^{\mathsf{T}}  \right)^{-1}\bm{H}^{(k+1)}\bm{M}^{(k)}
\end{align}

 and 
 
\begin{align}
        \bm{\beta}^{(k+1)} = \bm{\beta}^{(k)} + \bm{M}^{(k+1)}\left(\bm{H}^{(k+1)}\right)^{\mathsf{T}} \notag \\ \left(\bm{\theta}^{(k+1)}-\bm{H}^{(k+1)}\bm{\beta}^{(k)}\right)
\end{align}

\begin{figure}[h]
\centering
\vspace{-0.1cm}
\includegraphics[width=3.4in]{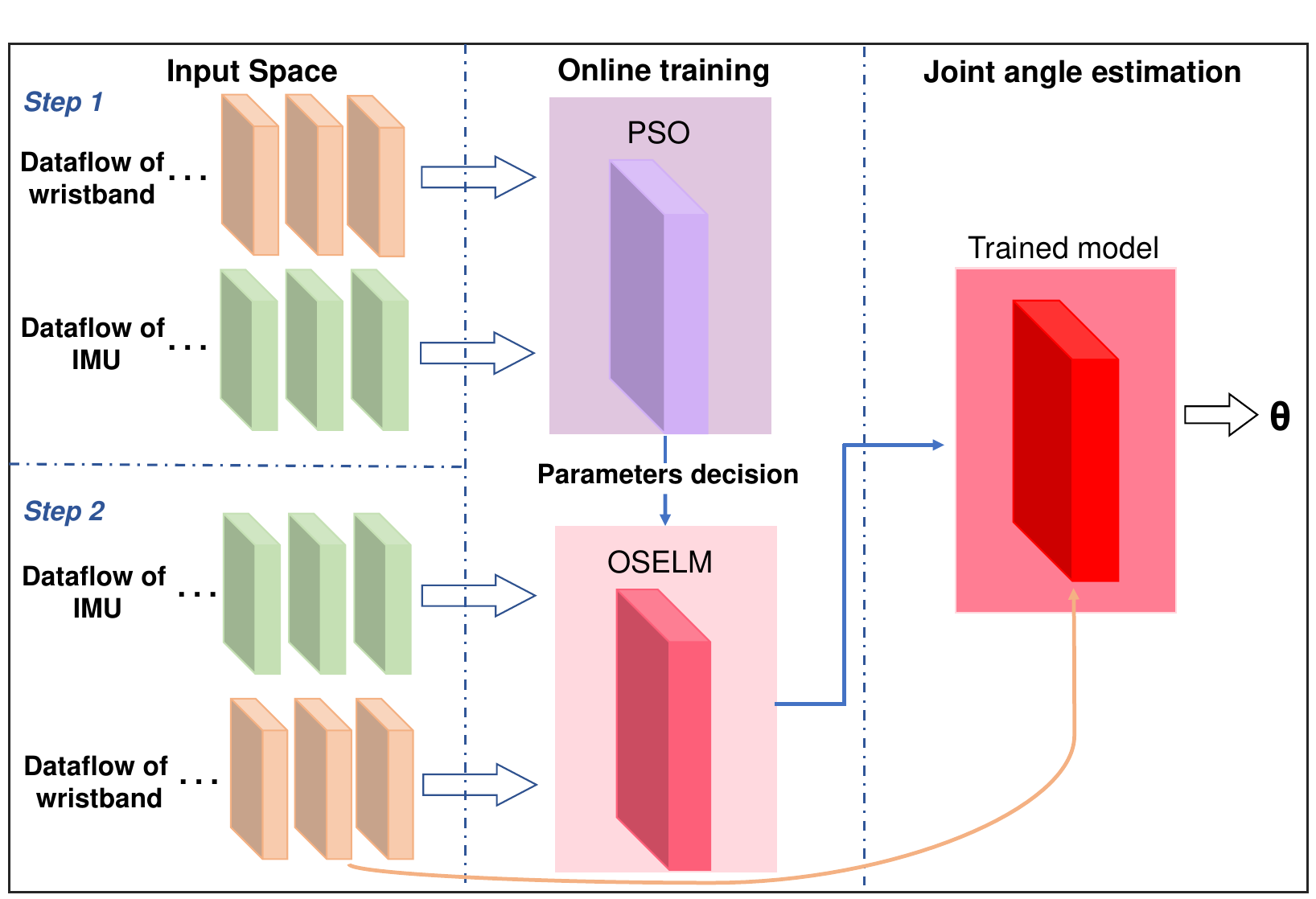}
\caption{Framework and workflow of the proposed method.}
\label{Framework}
\vspace{-0.6cm}
\end{figure}

After a certain number of samples are processed, the model can characterize the wearer's wrist movement patterns, and then online joint angle estimation can be carried out. The whole framework and workflow of the proposed method are visualized in \cref{Framework}.       

\
\
\section{EXPERIMENTS}
\subsection{Experiment Setup and Process}
\label{Experiment Setup and Process}
Two serial ports are utilized to read the dataflow of the wristband and IMU. At the same time, dataflow will be recorded, and the proposed model will be updated online. For wristband and IMU, the baud rates are 9600 and 115200. There are four strain sensors on the wristband, and the time interval between readings from adjacent sensors is 50 milliseconds. Compared to wristband, the angles information from IMU is much faster, for each complete readings of four strain sensors, about six to seven group of angles information will be captured. The data alignment method will be demonstrated in section \ref{Online Data Alignment}. The IMU is attached to the back of the hand, with its placement referenced to the middle finger, and its orientation $\theta{x}$, $\theta{y}$ and $\theta{z}$ vary with wrist motion.

In order to better replicate the experiments and evaluate the results, this work primarily focuses on the orientation $\theta_{y}$ for experiments and analyses based on the placement of IMU. Two different kinds of tests were performed on a subject. \cref{Wrist motion} below shows the wrist motion in this work.
\begin{figure}[h]
\centering
\vspace{-0.3cm}
\includegraphics[width=3.4in]{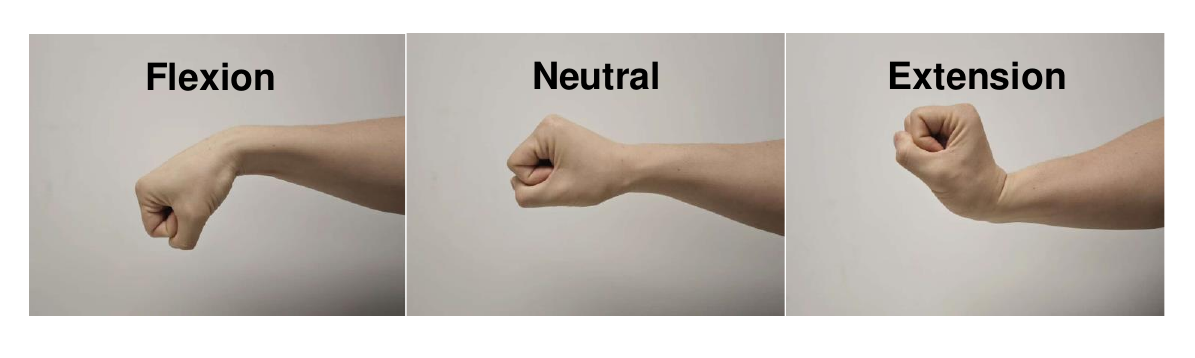}
\caption{Wrist motion.}
\label{Wrist motion}
\vspace{-0.2cm}
\end{figure}

\textit{1) The First Experiment}: In the first experiment, subject wears wristbands on his left or right wrist, and performs flexion and extension movements with the upper arm stationary on the support for about 3 minutes, which can be seen in \cref{Test Results with Wristband Misalignment.}(a), (b) and (c). At the beginning of the experiment, 10 seconds of the wearer's random hand rotational movements are recorded to calculate the parameters of ELM with PSO. After calculation, the wearer will perform cycles of wrist flexion and extension movements. A cycle is defined as the sequential movement from flexion to a neutral position, followed by extension, and concluding with a return to the flexion position. The speed of the wearer's wrist movement is almost uniform, which is about 5 to 10 seconds for a complete cycle. Then, while the subject is performing the defined actions, the model will update at the same time. The first 25\% data is used for online model updating, and part of the rest data will be tested to visualize the results of the proposed model. In this experiment, the program automatically differentiates between the PSO computation step and the online model training step, thus achieving process automation. To evaluate the effectiveness of the proposed method across diverse objects, participant will be instructed to complete the aforementioned procedure using their non-dominant hand, followed by performance assessment. The wristband will be positioned as consistently as possible on both hands. Given the differences in movement patterns and force exertion between an individual's dominant and non-dominant hands \cite{kim2023analysis, noguchi2009superiority}, including variations in motor habits, testing on both wrists allows for demonstrating the method's sensitivity to these differences.

\textit{2) The Second Experiment}: In the second experiment, to mimic the unavoidable placement variations encountered across distinct wearings, a distinct sensor offset is artificially applied to the wristband and wrist, and it should be noted that this work posits that the wristband remains largely stable throughout a given usage period, with deviations primarily arising between independent applications. Then, the aforementioned online training and calculation process will be done to evaluate the performance of the method. It is noteworthy that during the training phase, data from both the IMU and the strain sensors are utilized. However, following the training phase, angle estimation relies solely on the strain sensors' data, and the IMU is exclusively employed to acquire ground truth values and is not included in the operational implementation. 
 
\subsection{Online Data Alignment}
\label{Online Data Alignment}
As has mentioned in section \ref{Experiment Setup and Process}, for each set of strain readings, six to seven corresponding sets of angular measurements are acquired. A data queue is implemented for the storage and analysis of these two distinct data types. Because the IMU's reading rate is several times faster than that of the strain sensor, this paper uses the strain sensor readings as a time base. It assumes that subsequent sets of angle measurements acquired are temporally aligned with the moment of that particular strain sensor reading, until the next strain sensor reading is obtained. Furthermore, due to the slow movement speeds of the subject during the experiment, the several sets of angle measurements acquired between each strain reading are averaged, and this average value is regarded as the angle corresponding to the strain sensors readings at that moment.

\subsection{Performance Evaluation}
Because the subject primarily performed wrist flexion and extension movements, and given the IMU's placement on the hand, the IMU primarily measured the cahnging values of $\theta{y}$. Therefore, the evaluation of results will primarily be based on the estimation of $\theta{y}$.

The coefficient of determination ($R^2$) is utilized to measure the goodness of alignment between the fitting data and the ground truth provided by the IMU, which is defined as
\begin{equation}
\label{deqn_ex14}
R^2 = 1 - \frac{\sum\limits_{i=1}^{n}({V_{i} - V_{i_{est}}})^2}{\sum\limits_{i=1}^{n}(V_{i} - V_{ave})^2}
\end{equation}
Where $n$ is the number of samples, $V_{i}$ is the ground truth of the $i_{th}$ test point, $V_{i_{est}}$ is the corresponding estimation and $V_{ave}$ means the mean of $V_{i}$.

The mean error $Error_{ave}$ is used to measure the overall performance of the proposed method and sensor array.

\
\section{RESULTS}
\subsection{Results of Experiment One}
In this experiment, four strain sensors are distributed symmetrically on the wrist, with two sensors each on the dorsal and palmar sides, which will be shown in \cref{Test Results with Wristband Misalignment.}(a).
To visualize the estimation results of the proposed method, \cref{Test results of right and left wrist motion} below shows the test condition of right wrist and the trajectory changes of $\theta{y}$ (ground truth) provided by IMU and the corresponding calculation results obtained from the online trained model for both wrists. The $R^2$ value for right and left hand are 0.75 and 0.53, respectively.
For right wrist and left wrist joint angular estimation, the algorithm demonstrates good fitting performance, and the online-updated model effectively learns the motion patterns of both wrists at varying speeds. It can be noticed that when the speed of one wrist's movement is faster, it becomes more challenging for the fitting results to align with the IMU data at extreme maximum and minimum limits. This will be discussed in \cref{discussion}.
\begin{figure}[h]
\centering
\vspace{-0.1cm}
\includegraphics[width=3.4in]{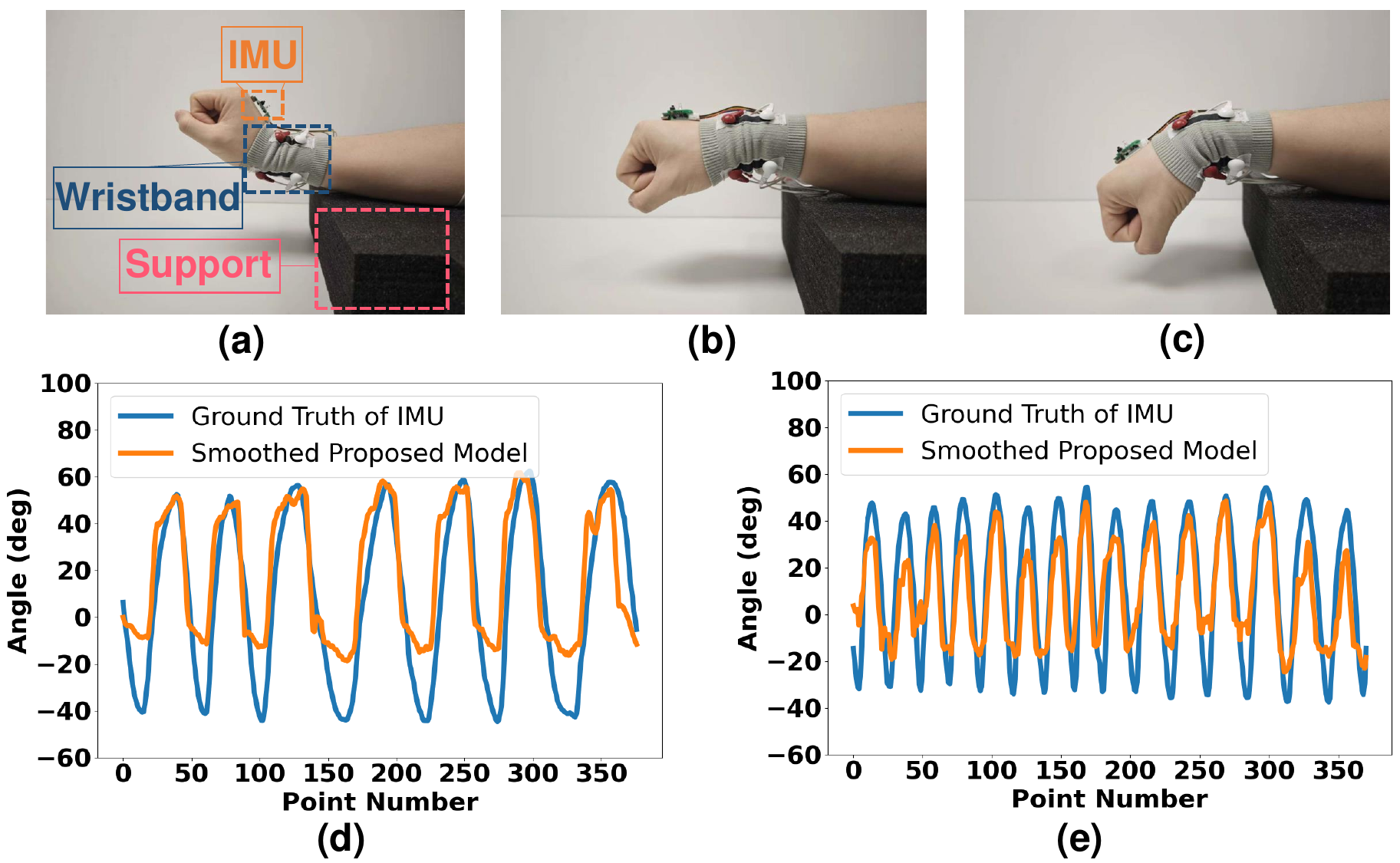}
\captionsetup{justification=justified, singlelinecheck=false}
\caption{Test results of right and left wrist motion. (a) Right wrist in a flexed position. (b) Right wrist in a neutral position. (c) Right wrist in an extended position. (d) Estimated trajectory of right wrist. (e) Estimated trajectory of left wrist.}
\label{Test results of right and left wrist motion}
\vspace{-0.1cm}
\end{figure}

\subsection{Results of Experiment Two}
\label{Results of Experiment Two}
\cref{Test Results with Wristband Misalignment.} below illustrates the effectiveness of the proposed method when there is a noticeable offset in the wearing position. \cref{Test Results with Wristband Misalignment.}(a) is the same result as \cref{Test results of right and left wrist motion}(d), which is the result of right wrist with four sensors uniform distributed on the dorsal and palmar side. As observed, even with significant sensor drift, this online learning method is effective for wrist joint angle estimation. Due to the unique features of each strain sensor, the proposed method can learn the wrist's motion patterns to some extent. However, the final results are still influenced by the inherent performance of the sensors, which can be noticed from the differences in calculation of maximum and minimum values. The general trajectory trend is consistent with the ground truth, and the $R^2$ value is 0.58 under this condition, which is mainly influenced by the calculation error of the maximum and minimum values.

\begin{figure}[h]
\centering
\vspace{-0.1cm}
\includegraphics[width=3.4in]{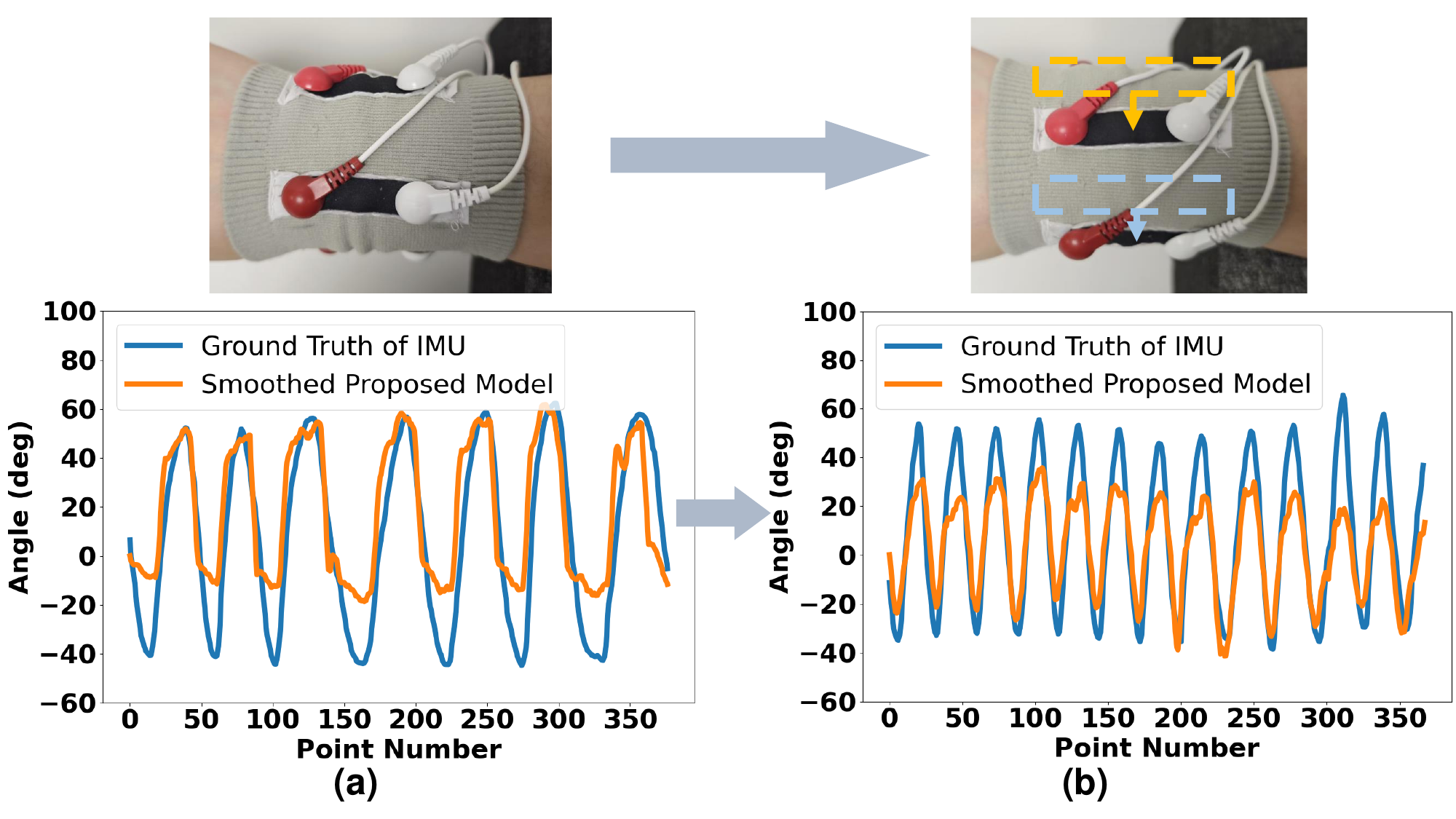}
\captionsetup{justification=justified, singlelinecheck=false}
\caption{Test results with wristband misalignment. (a) Test results for wristband in position one. (b) Test results for wristband in position two.}
\label{Test Results with Wristband Misalignment.}
\vspace{-0.1cm}
\end{figure}

\subsection{Discussion}
\label{discussion}
As can be observed from \cref{tab: Average error} below, the result in each column correspond to \cref{Test results of right and left wrist motion}(d), (e) and \cref{Test Results with Wristband Misalignment.}(b), respectively. Overall, the proposed system, through models trained with brief online learning, can track the flexion and extension movements of the wearer's wrist in various experimental scenarios. However, errors persist in the calculation of upper and lower bounds, which may be related to the sensor's inherent performance characteristics, such as measurement limits, or to sensor consistency differences leading to some sensors being more sensitive to strain in a particular direction. Furthermore, the online data alignment method used may still require further refinement to accommodate faster wrist movements, although it is sufficient for testing requirements under most of usage conditions. In addition, observing the average results in \cref{tab: Average error}, we can indicate that the calculated error remains within a close range across different scenarios, demonstrating stability in tracking motion trajectories. Nevertheless, deviations in the calculation of maximum and minimum values may still occur due to sensor variability or time-lag effects caused by the inability of the fabric to quickly return to a stable state after stretching, which is an issue that needs to be addressed in future work. Finally, observation of the resulting curves reveals that the calculation results do not always achieve good performance in estimating the initial and final actions. This is due to sensor characteristics, which means after being in a stretched state for a period of time, the sensor values slowly decrease. Therefore, it can be considered that it has better performance in tracking dynamic processes.

\begin{table}[h]
\centering
\captionsetup{justification=centering}
\caption{Average error of the experiments}
\label{tab: Average error}
\begin{tabularx}{\columnwidth}{XXXc}
\toprule
 & \multirow{3}*{\makecell{Right \\ wrist}} & \multirow{3}*{\makecell{Left \\ wrist}} & \multirow{3}*{\makecell{Right wrist with \\ sensor misalignment}}  \\
 & & & \\
 & & & \\
\midrule
\multirow{3}*{\makecell{$Error_{ave}$ \\(deg)}}  &  \multirow{3}*{\makecell{14.8}} & \multirow{3}*{\makecell{16.0}} & \multirow{3}*{\makecell{16.1}}   \\
& & & \\
& & & \\
\bottomrule
\end{tabularx}
\end{table}

\section{CONCLUSION}
This work first proposes to combine the advantages of PSO and OSELM to realize the online training framework for wrist joint angle estimation with customized fabric wristband, and it is the first evaluation of this framework under different test configurations. The results show that this framework has acceptable performance in wrist motion trajectory calculation with a $R^2$ value more than 0.7 under certain configuration. However, further improvements are needed in the consistency and stability of the proposed sensor array for fitting of angle extremes, and the multi-degree-of-freedom experiments on different users should be carried out. Overall, this framework is more practical and closer to real-world scenarios compared with other methods, and the experimental results demonstrate its feasibility.





\small
\bibliographystyle{unsrt}

\bibliography{iros_ref}  

\end{document}